\documentclass[journal]{IEEEtran}
\ifCLASSINFOpdf
\else
\fi
\usepackage{graphicx}
\usepackage{multirow}
\usepackage{mathrsfs}
\usepackage{makecell}
\usepackage[sort,nocompress]{cite}
\usepackage{amsfonts}
\usepackage{amssymb}
\usepackage{amsmath}
\usepackage{subcaption}
\usepackage{url}
\usepackage{booktabs}       
\usepackage{diagbox}
\usepackage{wrapfig}
\usepackage[table,xcdraw]{xcolor}
\usepackage[colorlinks=true, linkcolor=blue, citecolor=blue, urlcolor=blue]{hyperref}
\begin{document}
%

\title{FreqPKC: Frequency-Aware PeakConv for \\
    Radar Semantic Segmentation}
%
%
%


\author{Yangyong Deng,
    Pengyu Zhang,
    Shaoqiu Song,
    Xiaotao Huang, \textit{Senior Member, IEEE},
    and Jian Wang
    \thanks{The authors are with the College of Electronic Science and
        Technology, National University of Defense Technology, Changsha, Hunan
        410073, China. (e-mail: dengyy@nudt.edu.cn; zhangpengyu19@nudt.edu.cn;
        songshaoqiu@nudt.edu.cn; xthuang@nudt.edu.cn;
        jianwang\_uwb@nudt.edu.cn).}
    \thanks{Yangyong Deng and Pengyu Zhang contributed equally to this work.}
    \thanks{Corresponding to Jian Wang e-mail: jianwang\_uwb@nudt.edu.cn.}
}

\maketitle

\begin{abstract}
    Automotive millimeter-wave (mmWave) radar has attracted increasing attention
    for semantic segmentation in autonomous driving, owing to its robustness
    under adverse weather and lighting conditions. Recent works have introduced
    radar-specific neural operators such as PeakConv (PKC), inspired by constant
    false alarm rate (CFAR) detection, to better model the peak-dominated
    characteristics of radar images. However, existing methods typically treat
    the guard bandwidth as learnable parameters, without explicitly
    characterizing the frequency response of the PKC operator. In this work, we
    revisit PKC from a frequency-domain perspective and propose Frequency-Aware
    PeakConv (FreqPKC), which explicitly models and exploits its underlying
    spectral characteristics. By analyzing the frequency response of PKC,
    FreqPKC enables the network to capture a broader range of frequency
    components and adaptively adjust its effective bandwidth. With the proposed
    Frequency-Aware (FA) module, the network can infer appropriate guard
    bandwidths to balance high- and low-frequency information: it enhances
    high-frequency components in target regions for more precise segmentation
    while suppressing noise through low-frequency filtering in background areas.
    Building on FreqPKC, we further develop FreqPKC-Net, a radar semantic
    segmentation (RSS) framework that integrates frequency-guided adaptive
    modulation with cross-view fusion for multi-view inputs, and can also be
    adapted to single-view data. Experiments on the public datasets CARRADA,
    CARRADA-RAC, and a self-built dataset named RS-SAR demonstrate that
    FreqPKC-Net consistently outperforms existing SoTA methods on both
    range-Doppler and range-angle views.
\end{abstract}


%
\IEEEpeerreviewmaketitle

\section{Introduction}

\begin{figure}[!t]
    \centering
    \begin{subfigure}[b]{0.46\linewidth}
        \centering
        \includegraphics[width=\linewidth]{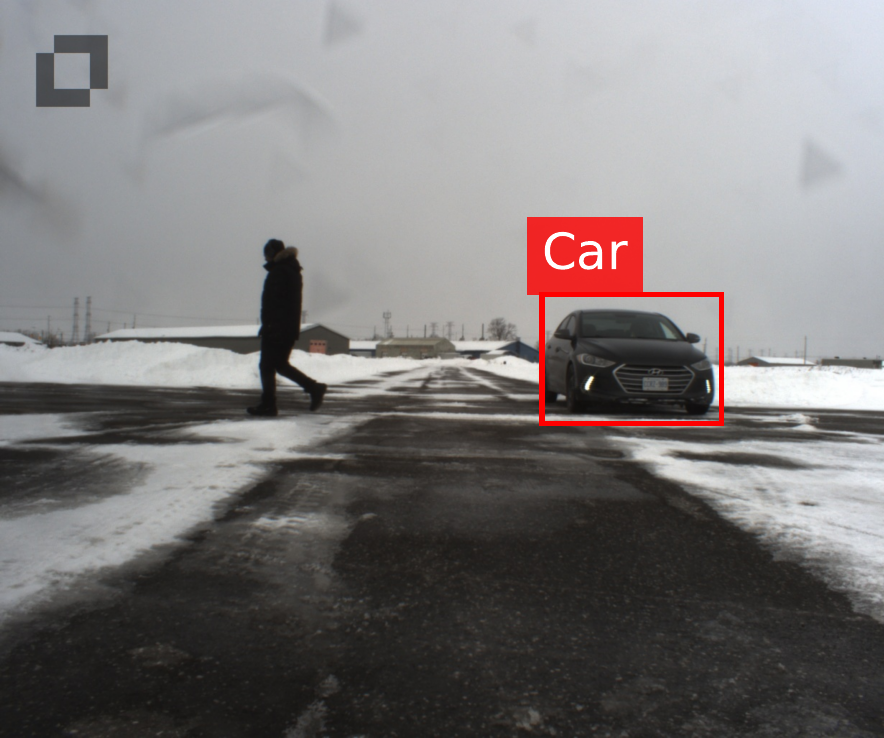}
        \caption{Camera Image}
        \label{fig:intro_cam}
    \end{subfigure}
    \hfill
    \begin{subfigure}[b]{0.46\linewidth}
        \centering
        \includegraphics[width=\linewidth]{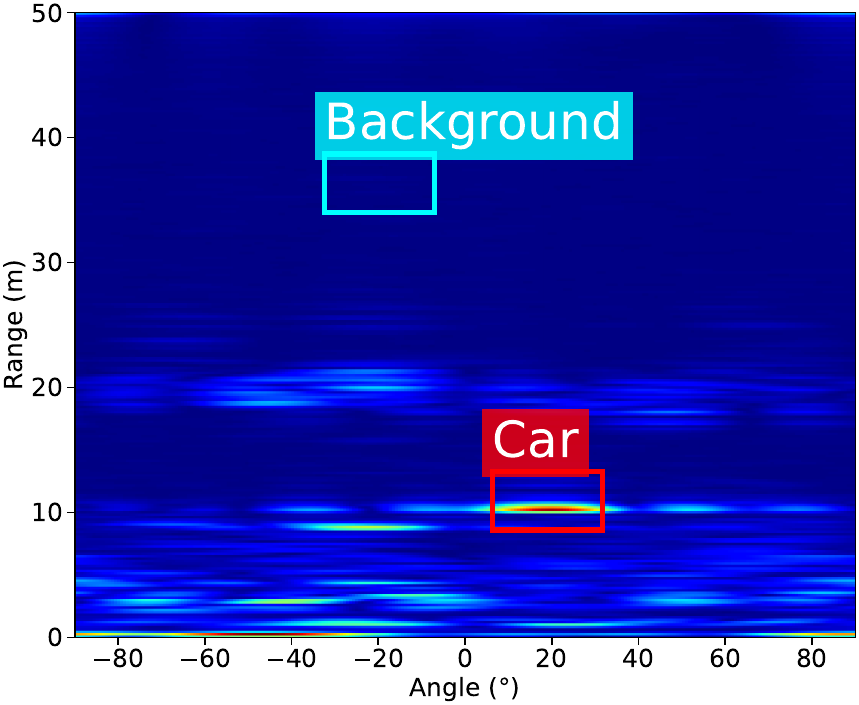}
        \caption{Range-Angle Map}
        \label{fig:intro_radar}
    \end{subfigure}

    \begin{subfigure}[b]{0.50\linewidth}
        \centering
        \includegraphics[width=\linewidth]{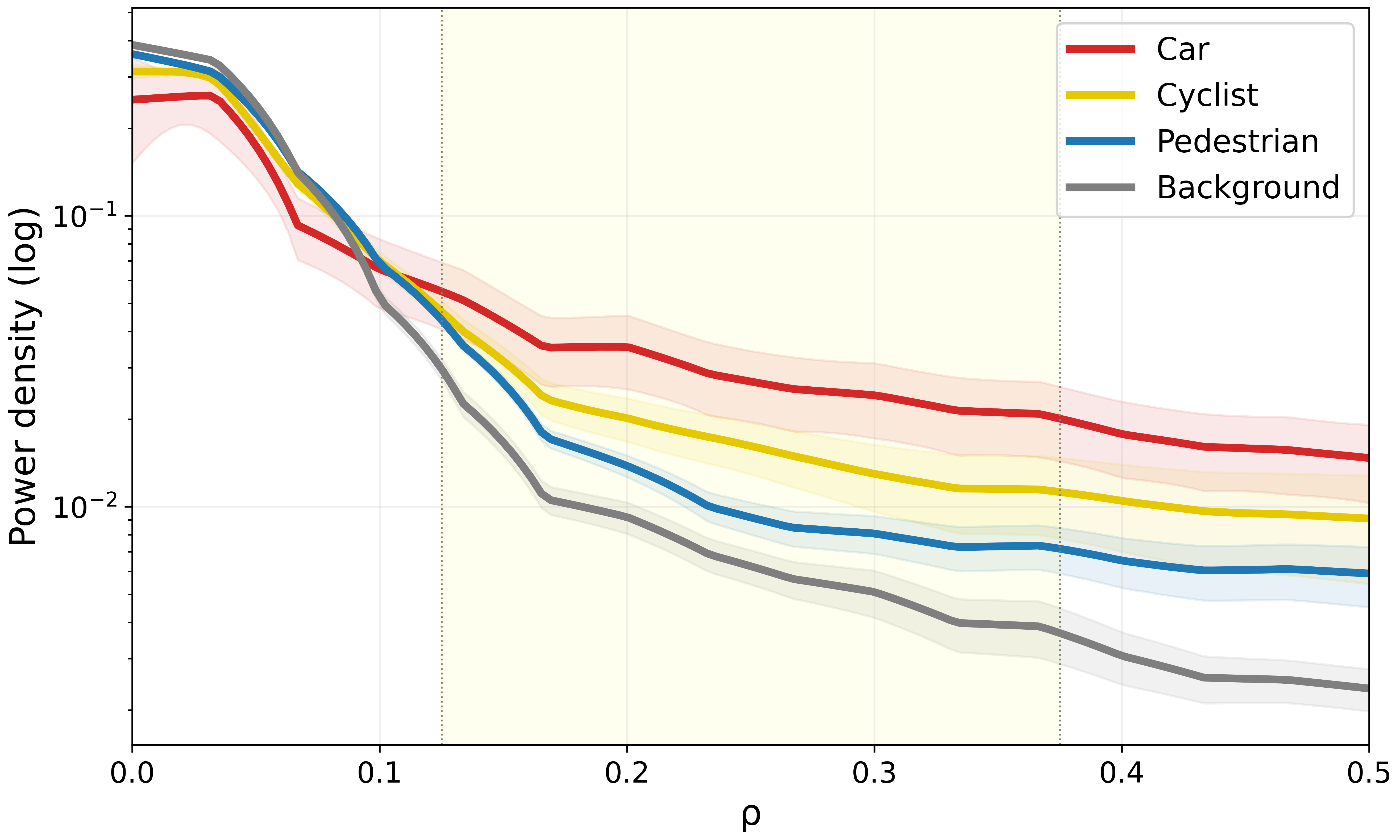}
        \caption{Class-wise Mean Spectrum}
        \label{fig:intro_class_spectrum}
    \end{subfigure}
    \hfill
    \begin{subfigure}[b]{0.46\linewidth}
        \centering
        \includegraphics[width=\linewidth]{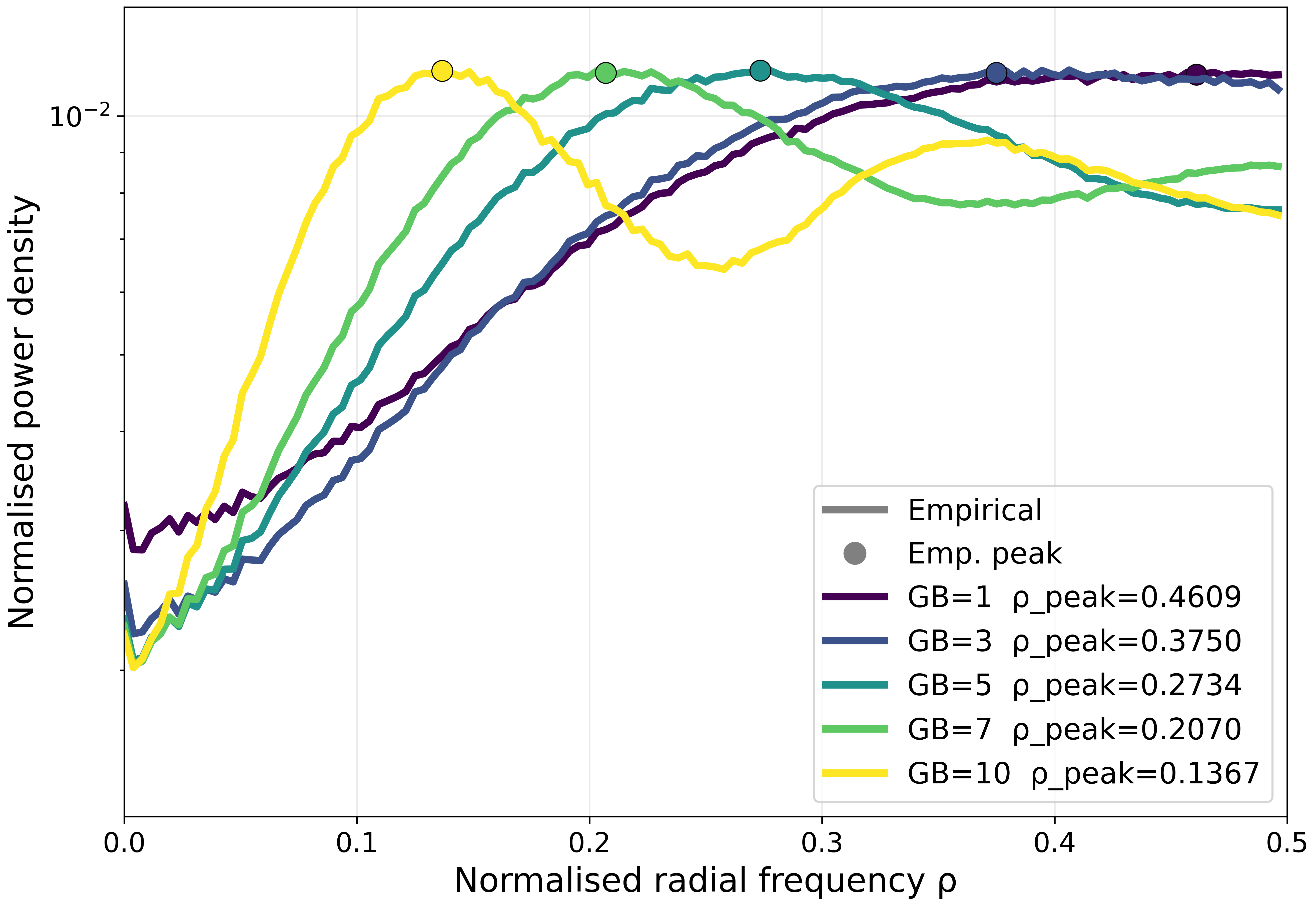}
        \caption{PKC Frequency Response}
        \label{fig:intro_freq_response}
    \end{subfigure}

    \caption{Data processing and spectral analysis visualization. (a) Optical camera
        image of a driving scene. (b) Corresponding Range-Angle (RA) radar map.
        (c) Class-wise normalized mean power spectra computed over the entire
        CARRADA-RAC
        dataset, showing the spectral separability between different semantic
        categories (Car, Pedestrian, Cyclist, and Background). The yellow band
        highlights the mid-frequency region $\rho \in [1/8, 3/8]$ where
        target-background discrimination is most pronounced. (d) Measured
        radial power spectra of the PKC operator under white-noise excitation
        for varying guard bandwidths $b_G$, validating the bandpass nature of
        PKC and the monotonic decrease of the passband center
        $\rho_{\rm peak}$ with $b_G$.}
    \label{fig:intro_spectrum}
\end{figure}
The rapid advancement of autonomous driving technology imposes stringent
requirements on the all-weather robustness of environmental perception systems \cite{zhang2026velocity, zhang2025seqlgpr, yin2025indoor, li2025seamless}.
Under adverse conditions such as rain, fog, and extreme lighting, mainstream
sensors such as cameras and LiDAR often suffer from performance degradation
\cite{sheeny2021radiate}. In contrast, mmWave radar has become an essential
sensor for perception systems due to its low cost and robustness to weather and
lighting variations \cite{patole2017automotive, sun2020mimo, song2024efficient, song2024efficient_image, jin2024srdst, wu2023novel}.
Despite these advantages, fully leveraging
the sensing capability of radar in complex and dynamic driving scenarios
necessitates a higher level of scene understanding. To this end, radar semantic
segmentation (RSS) has emerged as a key task.
RSS assigns pixel-level semantic
labels to radar tensors, enabling the perception system to recover structured
scene semantics from sparse radar measurements and thereby support downstream
planning, decision-making, and other related tasks \cite{madani2022radatron,
    ouaknine2021carrada, kaul2020rss, abdu2021application}.

Motivated by the success of convolutional neural networks (CNNs) in vision tasks
\cite{chen2017rethinking, chen2018encoder, he2017mask, yu2015multi, long2015fully}, early radar
perception methods also adopted standard CNN architectures
\cite{brodeski2019deep, kaul2020rss, ouaknine2021multi, major2019vehicle}.
Although this design choice seems intuitive, it introduces a clear domain gap.
Standard convolutions are built for dense RGB images, whereas radar measurements
are sparse and typically dominated by localized peak responses
\cite{yao2023radar, deng2025advances}.
Consequently, conventional CNNs waste computation on empty regions and tend to smooth out sharp target peaks.

To address the gap, Zhang et al. \cite{zhang2023peakconv} introduced the concept of the peak receptive field (PRF), inspired by constant false alarm rate (CFAR)
detection theory. They proposed PeakConv (PKC), which uses guard bandwidth to
preserve target peaks and suppress self-masking effects. In CFAR, the guard
bandwidth prevents target energy leakage from the cell under test (CUT) into
reference cells \cite{richards2005fundamentals, rohling1983radar,
    jalil2016analysis, trunk1978range}.

PKC relies on a fixed guard bandwidth, and subsequent adaptive variants
\cite{li2024adapkc} simply treat it as a learnable parameter without
accounting for the physical characteristics behind radar images. Although PKC
enhances the interpretability of neural-network-based radar image processing,
its potential remains underexplored.
Inspired by frequency-domain processing in traditional radar signal processing,
we further investigate the connection between the PRF and frequency response. We
first explore the frequency characteristics of target structures versus
background clutter in mmWave radar images, as shown in Fig.~\ref{fig:intro_spectrum}. We
select a representative sample from the CARRADA dataset
\cite{ouaknine2021carrada} to compare the optical image with the range-angle
(RA) map, and examine the class-wise spectral energy distribution computed
over the entire CARRADA-RAC dataset. The normalized power spectrum in
Fig.~\ref{fig:intro_spectrum}(\subref{fig:intro_class_spectrum}) shows that targets and background exhibit
clearly distinct frequency profiles: background clutter (gray solid line) is
predominantly low-pass, with energy decaying rapidly at higher frequencies,
whereas strong-scattering targets such as vehicles (red solid line) retain
substantial high-frequency components, forming a long-tail distribution. This
difference is consistent with their physical properties: target regions often
contain sharp spatial transitions and complex scattering structures, manifesting
as high-frequency energy, while background clutter is relatively smooth and
concentrated at low frequencies. As derived in Section~\ref{sec:theory}, the PKC
operator acts as a tunable bandpass filter whose passband center scales as
$1/b_G$. A smaller guard bandwidth ($b_G = 1$ in
Fig.~\ref{fig:intro_spectrum}(\subref{fig:intro_freq_response})) shifts the passband toward higher
frequencies, selectively capturing the sharp spatial transitions and scattering
centers characteristic of vehicle targets. Conversely, a larger guard bandwidth
($b_G = 3$) shifts the passband toward lower frequencies,
providing broader reference context that suppresses the low-pass-dominated
background clutter. This reveals that the guard bandwidth directly governs the
frequency selectivity of the PKC operator, and that spectral features serve as
a physically grounded bridge between target structure and adaptive PRF
modulation.



To address these limitations, we propose \textbf{Freq}uency-Aware \textbf{P}ea\textbf{kC}onv (FreqPKC), a radar-specific neural operator that adaptively modulates the peak receptive field using spectral energy distributions.
Unlike prior methods that rely on fixed or blindly learned guard bandwidths, FreqPKC explicitly leverages frequency-domain information to guide adaptive PRF modulation.
Central to our design is a three-module cascade: frequency decoupling, band-level learnable feature modulation, and frequency-aware modulated guard bandwidth. This cascade forms a fully differentiable operator that translates global spectral context into local sampling strategies in a physically interpretable way. Building on this operator, we construct FreqPKC-Net, which takes range-Doppler (RD) and range-angle (RA) tensors (with the angle-Doppler view as an auxiliary input) and outputs pixel-wise semantic segmentations for both representations simultaneously.

Our main contributions are summarized as follows:
\begin{itemize}
    \item We reveal that the guard bandwidth in PKC directly governs the
          frequency passband of the operator, establishing a principled
          connection between spectral characteristics and PRF design.

    \item Building on this insight, we propose Frequency-Aware PeakConv
          (FreqPKC), a radar-specific neural operator that uses cascaded
          frequency decoupling, band-level learnable modulation, and
          spectral-energy-driven guard bandwidth estimation to achieve
          physically grounded adaptive PRF control.

    \item We construct the Vehicle-Mounted Millimeter-Wave \textbf{R}otary
          \textbf{S}can \textbf{S}ynthetic \textbf{A}perture \textbf{R}adar
          (\textbf{RS-SAR}) Image Dataset, captured by a Rotary Scan SAR imaging
          system. Rotary Scan SAR imaging achieves an angular resolution better than
          0.8$^\circ$~\cite{song2026jsaf, zhu2025pseudo}, substantially finer than that of conventional automotive
          radars. RS-SAR also provides full 360$^\circ$ panoramic coverage,
          complementing the forward-facing FMCW radars in existing datasets.
\end{itemize}

The remainder of this paper is organized as follows.
Section~\ref{sec:related} reviews related work on automotive radar perception, radar semantic segmentation, and the peak receptive field.
Section~\ref{sec:method} presents the proposed FreqPKC operator, detailing the frequency decoupling, band-level modulation, and frequency-aware guard bandwidth estimation, followed by the overall FreqPKC-Net architecture with cross-view fusion.
Section~\ref{sec:experiments} describes the experimental setup, including datasets and training settings, and provides comprehensive evaluations covering frequency selectivity analysis, frequency band selection, comparison with SoTA methods on CARRADA and CARRADA-RAC, and results on the RS-SAR dataset.
Section~\ref{sec:conclusion} concludes the paper with a summary of key findings and future research directions.

\section{Related Work}
\label{sec:related}

This section reviews three lines of work most relevant to FreqPKC: automotive
radar perception, radar semantic segmentation, and the peak receptive field.
For each, we highlight the limitations that motivate the proposed
frequency-aware design.

\subsection{Automotive Radar Perception}
Automotive radar perception methods can be broadly divided into point
cloud-based and heatmap-based paradigms. Point cloud-based approaches operate on
sparse radar point clouds generated by on-board signal processing pipelines.
Representative methods include RPFA-Net~\cite{xu2021rpfa}, which employs
pillar-based feature attention for 3D object detection from radar point clouds,
and RadarPillars~\cite{musiat2024radarpillars}, which introduces velocity-aware
pillar representations for efficient 4D radar perception; LiDAR-oriented
detectors~\cite{lang2019pointpillars,yan2018second,shi2020pv,yin2021center}
have also been adapted to radar point clouds. Although these methods demonstrate
the feasibility of point cloud-based radar perception, the conversion from raw
radar signals to point clouds inevitably introduces substantial information
loss. Furthermore, the resulting sparse representation limits spatial resolution
and localization accuracy in downstream perception
tasks~\cite{madani2022radatron,yao2023radar,yao2025exploring}.

Heatmap-based methods operate directly on structured radar tensor
representations, such as range-Doppler (RD), range-angle (RA), and range-angle-Doppler (RAD)
tensors, thereby preserving the full spectral content of reflected signals.
RODNet~\cite{wang2021rodnet} proposes a cross-modal supervision strategy for
radar object detection directly from range-azimuth heatmaps.
FFT-RadNet~\cite{rebut2022raw} estimates object angles directly from
range-Doppler spectra, avoiding the explicit computation of the full RAD tensor.
It further performs object detection and free-space segmentation in a unified
framework. RADDet~\cite{zhang2021raddet} further exploits 4D radar tensors for
end-to-end object detection, demonstrating the effectiveness of structured radar
representations for perception tasks. Compared with point clouds, radar heatmaps
preserve richer scattering and fine-grained spectral structures, and exhibit a
distinct sparse peak-response property: energy concentrates around strong
scatterers while the background is dominated by diffuse
clutter~\cite{madani2022radatron,yao2023radar,yao2025exploring}. This
motivates task-specific operators rather than direct adaptation of operators
designed for dense visual imagery. In this work, we focus on heatmap-based RSS,
which provides pixel-level semantic understanding of radar tensors for
autonomous driving scene understanding.

\subsection{Radar Semantic Segmentation}
RSS delivers fine-grained pixel-level predictions distinct from bounding-box
detection, offering superior scene understanding~\cite{zhang2025rass}. Early
works adapted vision architectures: the deep radar detector (DRD) integrated
segmentation into detection, outperforming cell-averaging CFAR
(CA-CFAR)~\cite{brodeski2019deep}; RSS-Net introduced an end-to-end framework with atrous
spatial pyramid pooling (ASPP) for multi-scale context~\cite{kaul2020rss}. To leverage the multi-dimensional nature of radar data, MVA-Net, TMVA-Net, and
RAMP-CNN jointly process RA and RD maps~\cite{ouaknine2021multi,gao2020ramp}.
However, these methods rely on naive
fusion and fail to consider cross-view spatial misalignment. AXFL addressed this
via axial-prior-guided alignment and adaptive fusion~\cite{li2025axfl}. For
temporal modeling, TMVA-Net used 3D CNNs with limited receptive fields; later,
T-RODNet and TransRSS incorporated Transformers for global context but faced
high computational
costs~\cite{jiang2022t,zou2023transrss,dalbah2024transradar}. TARSS-Net proposed
data-driven temporal aggregation to balance efficiency and
performance~\cite{zhang2024tarss}. Although these methods improve performance with
complex visual modules, they treat radar data as generic images, overlooking
the unique physical attributes of radar signals and causing computational
redundancy.

\begin{figure}
    \centering

    \begin{subfigure}[b]{0.44\linewidth}
        \centering
        \includegraphics[width=\linewidth]{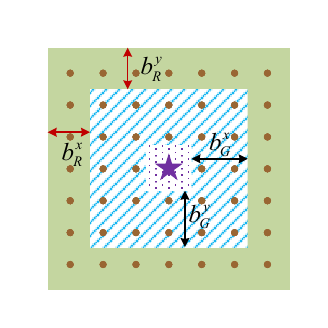}
        \caption{Guard/Reference Bandwidths}
    \end{subfigure}
    \hfill
    \begin{subfigure}[b]{0.44\linewidth}
        \centering
        \includegraphics[width=\linewidth]{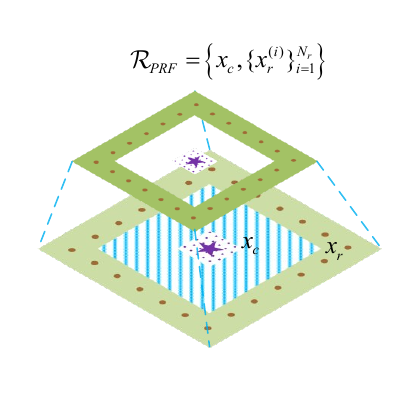}
        \caption{Peak Receptive Field}
    \end{subfigure}

    \caption{Schematic diagram of the Peak Receptive Field (PRF). For a center
    unit \(x_c\), the reference cells \(\{x_r^{(i)}\}_{i=1}^{N_r}\) are sampled
    outside the guard region defined by guard bandwidth \(\mathbf{b}_G\) and
    reference bandwidth \(\mathbf{b}_R\).}
    \label{fig:prf_schematic}
\end{figure}

\subsection{Peak Receptive Field}
The PRF was originally proposed by Zhang et al.~\cite{zhang2023peakconv}. Based
on this, PKC was established as the pioneering convolution paradigm explicitly
tailored for radar signal processing. Diverging from conventional operators,
\textit{e.g.}, dilated~\cite{yu2015multi} or deformable~\cite{dai2017deformable,zhu2019deformable} convolutions, PKC precisely captures the
peak features embedded in radar echoes by learning the PRF, thereby
demonstrating significant superiority in RSS tasks.

As illustrated in Fig.~\ref{fig:prf_schematic}, for any center unit $x_c$ on the
feature maps, its PRF is defined as a set composed of the center unit $x_c$
itself and a set of reference cells $\{x_r^{(i)}\}_{i=1}^{N_r}$. These reference
cells are distributed outside the guard region and are utilized to estimate
background interference. Given the guard bandwidth $\mathbf{b}_G = (b_G^x,
    b_G^y)$ and the reference bandwidth $\mathbf{b}_R = (b_R^x, b_R^y)$, the PRF
$\mathcal{R}_{PRF}$ is defined as:
\begin{equation}
    \mathcal{R}_{PRF} = \{x_c , \{x_r^{(i)}\}_{i=1}^{N_r}\}
\end{equation}
subject to the condition:
\begin{equation}
    |b_G| < |p_{r}^{(i)} - p_{c}| \leq |b_R + b_G|,
\end{equation}
where $p_{c}$ and $p_{r}^{(i)}$ denote the spatial coordinates of the center unit and the $i^{th}$ reference
unit, respectively.

However, the original PeakConv relies on a fixed PRF, making it difficult to
adapt to dynamic variations in target scale and clutter, and constraining the
network's learning potential. Subsequently, AdaPKC~\cite{li2024adapkc} achieved adaptive PRF
adjustment via two distinct strategies, \textit{i.e.}, AdaPKC$^\theta$, which
employs a lightweight network to estimate the optimal PRF, and AdaPKC$^\xi$,
which operates by comparing the similarity between the center cell and candidate
reference cells. Nevertheless, AdaPKC$^\theta$ sacrifices physical
interpretability to minimize the loss function, implying that the learned
bandwidths may not truly reflect signal fidelity. Meanwhile, AdaPKC$^\xi$ relies
heavily on the quality of spatial features, limiting its generalization
capability in complex electromagnetic environments. Both strategies, despite
their adaptivity, estimate the PRF solely from spatial-domain cues and overlook
a fundamental aspect of radar signals: target structures and background clutter
exhibit markedly different spectral signatures. In traditional radar signal
processing, frequency-domain analysis has long been a cornerstone for target
detection, yet this principle remains largely unexploited in PRF design. As a
result, existing adaptive methods operate in a physically agnostic manner,
treating the guard bandwidth as a generic learnable offset without any
frequency-selective guidance.

\section{Methodology}
\label{sec:method}

We first analyze the PKC operator's frequency response
(Section~\ref{sec:theory}), establishing the relationship between guard
bandwidth and spectral selectivity, then introduce the FreqPKC operator in
Section~\ref{sec:freqpkc} and the overall network in Section~\ref{sec:network}.

\begin{figure*}
    \centering
    \includegraphics[width=0.80\textwidth]{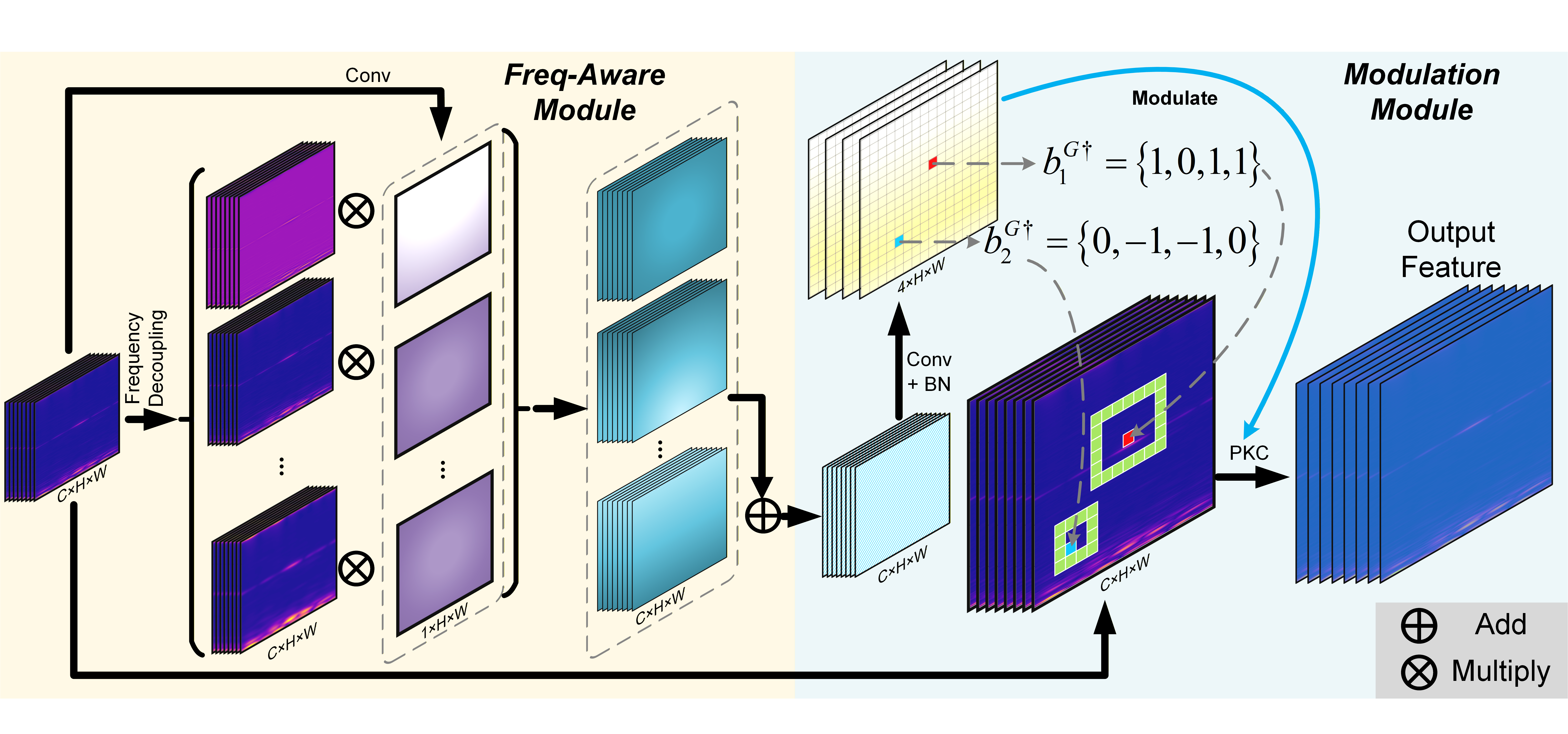}
    \caption{Overview of the Frequency-Aware PeakConv (FreqPKC). The input
    features are fed into the Frequency-Aware (FA) module for frequency
    decoupling, producing multiple sub-band features. Each sub-band is
    weighted and modulated by learnable weights (Multiply) and the modulated
    features are fused (Add); the Modulation Module then outputs the guard
    bandwidth offsets \(B^{G\dagger}\in \mathbb{R}^{4\times H\times W}\) that
    control the PKC operation, with \(b_1^{G\dagger}\) and \(b_2^{G\dagger}\)
    representing the offsets at two spatial locations, each specifying
    bandwidth adjustments in four directions (up, down, left, right).}
    \label{fig:freqpkc_pipeline}
\end{figure*}

\subsection{Theoretical Analysis of PeakConv}
\label{sec:theory}

\subsubsection{Frequency Response of the PKC Operator}

We model PKC as an equivalent linear shift-invariant kernel. Following the PRF
definition in~\cite{zhang2023peakconv}, the response-difference-aware
PKC (ReDA-PKC) operator directly learns
the response difference between the center unit $x_c$ and each reference cell
$x_r^{(i)}$:
\begin{equation}
    y_j = \sum_{i=1}^{N_r} \mathbf{w}_j^{(i)} \circledast (x_c - x_r^{(i)})
    = \underbrace{\Big(\sum_{i} \mathbf{w}_j^{(i)}\Big)}_{\mathbf{W}_c^{(j)}}
    \circledast x_c - \sum_{i=1}^{N_r} \mathbf{w}_j^{(i)} \circledast x_r^{(i)}
    \label{eq:pkc_reform}
\end{equation}
Equation~(\ref{eq:pkc_reform}) reveals a \emph{center-minus-surround}
architecture: the center weight $\mathbf{W}_c^{(j)}$, the sum of all reference
weights, acts on $x_c$, while the weighted sum of the reference responses is
subtracted from it. This arrangement is the spatial-domain signature of
CFAR-style interference cancellation.

For spectral analysis, we approximate the operator by an isotropic 2D kernel
with uniform scalar weights:
\begin{equation}
    K_{\rm PKC}(x, y; b_G, b_R) = w_c \cdot \delta(x, y) \;-\;
    w_r \cdot \mathbf{1}\!\left[b_G < r \leq b_G + b_R\right]
    \label{eq:pkc_kernel}
\end{equation}
where $r = \sqrt{x^2 + y^2}$, $b_G$ and $b_R$ are the guard and reference
bandwidths, and the coupling constraint $w_c = N_r \cdot w_r$ holds.
The minus sign before the annular reference region distinguishes PKC
from standard convolution's additive aggregation and is the origin of its
bandpass behavior.

Owing to the circular symmetry of $K_{\rm PKC}$, its 2D Fourier transform
reduces to the zero-order Hankel transform:
\begin{equation}
    \hat{K}_{\rm PKC}(\rho) = w_c - 2\pi w_r \int_{b_G}^{b_G+b_R} r\,J_0(2\pi\rho r)\,dr
    \label{eq:hankel}
\end{equation}
where $\rho = \sqrt{u^2+v^2}$ is the radial frequency and $J_0$ is the
zero-order Bessel function of the first kind. Using the identity
$\int rJ_0(ar)dr = \frac{r}{a}J_1(ar)$, we obtain the closed-form frequency
response:
\begin{equation}
    \begin{aligned}
        \hat{K}_{\rm PKC}(\rho) = w_c - \frac{w_r}{\rho}\,
            \big[(b_G+b_R)\,J_1(2\pi\rho(b_G+b_R)) \\
            -\, b_G\,J_1(2\pi\rho b_G)\big]
    \end{aligned}
    \label{eq:bessel}
\end{equation}
where $J_1$ is the first-order Bessel function.

Two properties follow from this closed form. First, with the small-argument
approximation $J_1(x) \approx x/2$ and the coupling constraint
$w_c = \pi w_r(2b_G b_R + b_R^2)$, the continuous-area counterpart of
$w_c = N_r w_r$ with $N_r \approx \pi(2b_G b_R + b_R^2)$, one obtains
$\hat{K}_{\rm PKC}(0) = 0$: the operator has zero DC gain and completely
suppresses the spatially uniform background component. This is the
frequency-domain manifestation of the CFAR principle: when the center cell
and its surroundings share the same intensity, the output is zero.

Second, the passband center and width follow from the first maximum of $J_1(x)$ at
$x \approx 1.841$ and its first two roots at $x \approx 3.832$ and $7.016$:
\begin{equation}
    \rho_{\rm peak} \approx \frac{1.841}{2\pi b_G} \propto \frac{1}{b_G},
    \qquad
    \Delta\rho_{-3{\rm dB}} \propto \frac{1}{b_G}
    \label{eq:scaling}
\end{equation}
This scaling is obtained for $b_R \gg b_G$, where the outer-edge Bessel term
oscillates rapidly and the passband is governed by the inner guard edge $b_G$.
The guard bandwidth $b_G$ therefore directly governs the passband position:
a smaller $b_G$ shifts the passband toward higher frequencies
to capture fine target structures, while a larger $b_G$ shifts it toward lower
frequencies for broad background reference estimation. This prediction is
empirically validated in Supplementary Fig.~S4, where the spatial distribution
of the effective guard bandwidth $b_{\rm eff} = 1 + {\rm mean}(B^{G\dagger})$
inferred by a trained model
confirms that target regions are consistently assigned smaller $b_G$ (higher
passbands) while background regions receive larger $b_G$ (lower passbands).

Fig.~\ref{fig:intro_spectrum}(\subref{fig:intro_freq_response}) validates
Eq.~(\ref{eq:bessel}) empirically: we feed i.i.d.\ white-noise inputs through
the PKC operator with five fixed guard bandwidths and radially average the
output power spectra. All measured spectra exhibit the bandpass shape
predicted by the theory, with the peak frequency decreasing monotonically as
$b_G$ increases and all curves converging to zero at DC, confirming the zero
DC gain and the tunability of the passband position.

\subsubsection{Frequency-Domain Estimation of the Guard Bandwidth}

Estimating $b_G$ solely from spatial features, as done by AdaPKC$^\theta$ and
AdaPKC$^\xi$~\cite{li2024adapkc}, suffers from an inherent ambiguity: a
high-intensity cell may be either a target edge or an isolated clutter spike,
and spatial intensity alone cannot disambiguate the two. Furthermore, the
black-box regression from spatial features to $b_G$ lacks physical constraints
and may converge to solutions that violate the radar signal model.

Frequency-domain features provide orthogonal discriminative information.
Radar targets and background exhibit distinct spectral signatures rooted in
electromagnetic physics: targets produce spatially localized but spectrally broad
scattering (sharp edges, multi-bounce, diffraction $\to$ mid-to-high spatial
frequencies), while background clutter is spatially smooth and predominantly
low-pass~\cite{richards2005fundamentals}. This spectral separation is
orthogonal to spatial intensity, providing a complementary discriminative
dimension that resolves the spatial ambiguity.

The proposed FA module (Section~\ref{sec:freqpkc}) exploits this complementarity
through cascaded frequency decoupling: it decomposes features into $M+1$
non-overlapping sub-bands, each encoding a distinct spectral scale. This
produces a \textit{spectral feature vector} at every spatial location, enabling
physically grounded $b_G$ decisions that purely spatial features cannot support.
The resulting frequency-guided modulation aligns the learned PRF with the
underlying spectral structure of radar signals, which is invariant to dataset-
specific biases and generalizes across different radar sensors (as validated by
the multi-dataset evaluations in Section~\ref{sec:experiments}).

\subsection{Frequency-Aware PeakConv}
\label{sec:freqpkc}

FreqPKC addresses the spatial-only estimation discussed above by explicitly
leveraging spectral energy distribution to guide adaptive guard bandwidth
modulation. As shown in Fig.~\ref{fig:freqpkc_pipeline}, a Frequency-Aware
(FA) module operates in parallel with peak convolution: it decomposes input
features into non-overlapping frequency bands, analyzes the energy statistics
of each band, and maps these spectral cues into directional guard bandwidth
offsets. Functioning as a parameter estimation network, it utilizes the global
spectral context of the radar signal to infer the optimal offset at each
spatial location, driving the subsequent peak convolution to perform
frequency-guided adaptive sampling on the original feature map.

\subsubsection{Frequency Decoupling Process}

\begin{figure}
    \centering
    \includegraphics[width=0.45\textwidth]{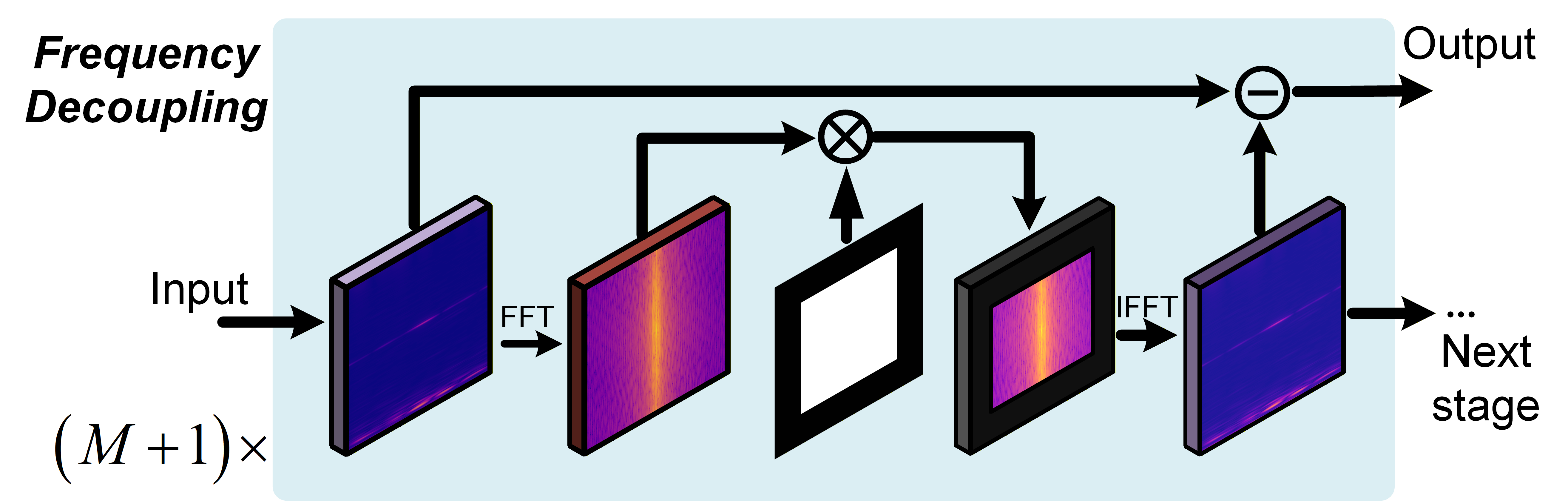}
    \caption{Schematic of the Frequency Decoupling process.}
    \label{fig:freq_decoupling}
\end{figure}

As shown in Fig.~\ref{fig:freq_decoupling}, given an input feature map $X \in \mathbb{R}^{C
        \times H \times W}$, we apply 2D FFT followed by a spectrum centralization
operation $Shift(\cdot)$ to relocate the zero-frequency component from the
spectrum corners to the geometric center, concentrating low-frequency energy
centrally to enable mask-based filtering:
\begin{equation}
    X_{freq} = Shift(\mathcal{F}(X))
    \label{eq:freq_shift}
\end{equation}
where $\mathcal{F}(\cdot)$ denotes the 2D FFT and the centralization allows
frequency masks to be defined via simple center cropping.

Different frequency components carry complementary semantic information, so we
adopt a cascading residual frequency decoupling strategy that divides features
into multiple non-redundant bands. To decouple different frequency components, we define a set of cutoff
frequency scale factors $\mathcal{K} = \{k_1, k_2, \ldots, k_M\}$, with $k_1
    < k_2 < \cdots < k_M$. For each $k_i \in \mathcal{K}$, we generate a binary
rectangular low-pass mask $M_{k_i}$ centered at the spectrum origin, which
retains only the central $\frac{H}{k_i} \times \frac{W}{k_i}$ region of the
centralized spectrum. A larger $k_i$ corresponds to a narrower low-frequency
passband, \textit{i.e.}, a lower effective cutoff frequency. Thanks to the spectrum
centralization operation, this mask can accurately cover and retain the
low-frequency components in the central region.

For the $i^{\text{th}}$ stage, we first extract the low-frequency part via
mask, then apply inverse centralization $Shift^{-1}(\cdot)$ and inverse
Fourier transform $\mathcal{F}^{-1}$ to restore it to the spatial domain. The
low-frequency component $X_{low}^{(i)}$ and the band detail component
$X_{band}^{(i)}$ are formulated as:
\begin{equation}
    X_{low}^{(i)} = Re\left(\mathcal{F}^{-1}\left(Shift^{-1}
        \left(X_{freq} \odot M_{k_i}\right)\right)\right)
\end{equation}
\begin{equation}
    X_{band}^{(i)} = X_{pre} - X_{low}^{(i)}
    \label{eq:band_residual}
\end{equation}
Initially, $X_{pre} = X$. After each decoupling stage, $X_{pre}$ is updated to
$X_{low}^{(i)}$ for the next iteration. This process yields a series
$\{X_{band}^{(i)}\}_{i=1}^{M+1}$, representing signal energy across distinct
frequency bands. The residual decoupling offers two advantages: 1) avoiding
information overlap between bands; 2) ensuring numerical stability and smooth
gradient propagation. Through cascaded residual subtraction (Eq.~\ref{eq:band_residual}),
$\mathcal{K}$ partitions the full spectrum into $M+1$ non-overlapping sub-bands
$\{X_{band}^{(i)}\}_{i=1}^{M+1}$, ordered from fine high-frequency target
scattering structures ($i=1$) down to the coarse low-frequency background
residual ($i=M+1$). The resulting
sub-bands are subsequently passed to the band-level modulation stage described
below.

\subsubsection{Band-level Learnable Feature Modulation}

For each decoupled sub-band $X_{band}^{(i)}$, a dedicated depthwise
convolutional layer whose weights are initialized to zero generates a spatial
attention map $A^{(i)}$ to derive a weight map for dynamic frequency
modulation:
\begin{equation}
    \tilde{X}_{band}^{(i)} = X_{band}^{(i)} \odot \sigma(A^{(i)})
\end{equation}
where $\odot$ denotes element-wise multiplication, and $\sigma(\cdot)$
represents the Sigmoid activation function. This band-independent design
explicitly injects frequency-correlated spatial priors into the network,
enabling the model to adaptively enhance frequency components associated with
target structures while suppressing bands dominated by noise or clutter.
Notably, the final low-frequency residual $X_{pre}$ is also modulated by an
additional attention map, so that all $M+1$ sub-bands contribute with learned
weighting; the re-weighted components are summed to yield the
frequency-modulated feature map $X_{weight}$.

Subsequently, an output head is employed to deduce the guard bandwidth offsets
$B^{G\dagger}$ for inferring the optimal PRF, utilizing a learnable mapping
function to project $X_{weight}$ into the guard bandwidth search space.
Specifically, a $(2b_{max}^G+1)\times(2b_{max}^G+1)$ convolutional layer
$\phi_{comb}$ followed by Batch Normalization (with $b_{max}^G=3$ in our
implementation, yielding a $7\times7$ kernel) fuses multi-band energy and
generates directional control parameters:
\begin{equation}
    \mathcal{A} = \phi_{comb}(X_{weight})
\end{equation}
where $\mathcal{A} \in \mathbb{R}^{4 \times H \times W}$ corresponds to the
attention weights for the four cardinal directions (\textit{i.e.}, up, down,
left, and right). Finally, the optimal guard bandwidth offset $B^{G\dagger}$ is
obtained via a Sigmoid activation function $\sigma(\cdot)$ scaled by the
maximum guard bandwidth $b_{max}^G$:
\begin{equation}
    B^{G\dagger} = (b_{max}^G - 1) \cdot \sigma(\mathcal{A})
\end{equation}
where $B^{G\dagger} \in \mathbb{R}^{4 \times H \times W}$ represents the
per-location guard bandwidth offsets in four directions, bounded within
$[0,\, b_{max}^G - 1]$. This mechanism empowers the network to discern which
frequency components contribute positively or negatively to the expansion of
the receptive field.

\subsubsection{Frequency-Aware Modulated Guard Bandwidth}

Upon acquiring the optimal guard bandwidth offset $B^{G\dagger}$, we proceed
to resample the reference cells $\{x_r\}$ on the feature map. Specifically,
$B^{G\dagger}$ is decomposed into four directional components, \textit{i.e.},
$B^{G\dagger} = \{b_{top}, b_{bottom}, b_{left}, b_{right}\}$. For every
convolution window centered at $p_c$, the initial fixed reference grid $p_r$
undergoes deformation driven by these components. The updated floating-point
coordinates $p_r'$ are computed as:
\begin{equation}
    p_r' = p_r + \mathcal{M}(B^{G\dagger})
\end{equation}
where $\mathcal{M}(\cdot)$ denotes the mapping function that projects the four
directional bandwidth scalars onto the coordinate offsets of each reference
point $(x, y)$. Given that the calculated $p_r'$ are typically non-integers
(\textit{i.e.}, sub-pixel coordinates) and cannot be directly indexed on a
discrete feature grid, we employ a differentiable linear interpolation
mechanism to compute the feature values. For an arbitrary floating-point
coordinate $p_r'$, we first locate its top-left integer neighbor $q_{lt} =
    \lfloor p_r' \rfloor$, and bilinear sampling aggregates the four integer
neighbors $q \in \{q_{lt},\, q_{lt}+(1,0),\, q_{lt}+(0,1),\, q_{lt}+(1,1)\}$.
The sampled value $x(p_r')$ is approximated via a weighted sum of neighborhood
features:
\begin{equation}
    x(p_r') = \sum_{q} G(p_r', q) \cdot x(q), \qquad
    G(p, q) = g(p_x, q_x)\, g(p_y, q_y)
\end{equation}
where $g(a, b) = \max(0,\, 1 - |a - b|)$ is the 1D linear interpolation
kernel applied along each axis, with $p_x, p_y$ and $q_x, q_y$ denoting the
coordinate components of $p$ and $q$. This sampling mechanism
not only resolves the coordinate discretization issue but, more crucially,
preserves differentiability with respect to the offset $B^{G\dagger}$. This
ensures that gradients of the guard bandwidth can be back-propagated to the FA
module via interpolation weights, thereby facilitating end-to-end training.

The FA module introduces only marginal computational overhead, with
lightweight FFT-based decoupling and a single $(2b_{max}^G{+}1)^2$
estimation-head convolution. A full efficiency comparison, including
parameters, GFLOPs, and FPS, is provided in the Supplementary Material
(Table~S1).

\subsection{The Overall Framework of FreqPKC-Net}
\label{sec:network}

Based on the FreqPKC operator, we propose FreqPKC-Net, an end-to-end multi-view
radar semantic segmentation network. As illustrated in the Supplementary
Material (Fig.~S1), it adopts a three-stream encoder and dual-stream decoder
architecture, where the AD branch serves as an auxiliary modality to enhance RD
and RA segmentation. Each view (RD, RA, AD) is processed by an identical
encoding branch consisting of a Double 3D Convolution Block for spatiotemporal
feature extraction, followed by two Double FreqPKC Blocks (each containing two
cascaded FreqPKC layers) where the FA module dynamically infers optimal guard
bandwidths. An ASPP module aggregates
multi-scale context after each branch. A Latent Space Encoder (LSE) fuses the
three view-specific latent embeddings, and two parallel decoders reconstruct
full-resolution segmentation masks for RD and RA views. The network is trained
end-to-end with a joint loss combining weighted cross-entropy, Dice, and
coherence terms on both decoder outputs. Full
architectural details are provided in the Supplementary Material.

\section{Experiments}
\label{sec:experiments}
To verify the effectiveness of the proposed frequency-aware method, we conducted
quantitative and qualitative experiments on two public multi-view radar datasets
and a self-built single-view radar dataset. To simplify notation, the multi-view
RSS model based on FreqPKC is denoted as FreqPKC-Net.

\subsection{Datasets and Training Setting}

We evaluate on three datasets. \textbf{CARRADA}~\cite{ouaknine2021carrada} provides
synchronized camera-radar data with three classes (car, pedestrian, cyclist) and
multi-view RAD tensors ($256 \times 256 \times 64$).
\textbf{CARRADA-RAC}~\cite{zhang2023peakconv} refines CARRADA's RA annotations.
\textbf{RS-SAR} is a self-built dataset captured by a 77\,GHz mmWave Rotary Scan
SAR system~\cite{hua2022panoramic}, as illustrated in Fig.~\ref{fig:radarsys} (see Supplementary
Material for further details), providing single-view RA images at $288 \times 288$
with two classes (small and large vehicles). Since only the RA view is
available, models evaluated on RS-SAR adopt single-view architectures (see
Supplementary Material). Standard train/val/test splits are
used following prior work~\cite{ouaknine2021multi,zhang2023peakconv,li2024adapkc}.

\begin{figure}[!t]
    \centering
    \includegraphics[width=0.72\linewidth]{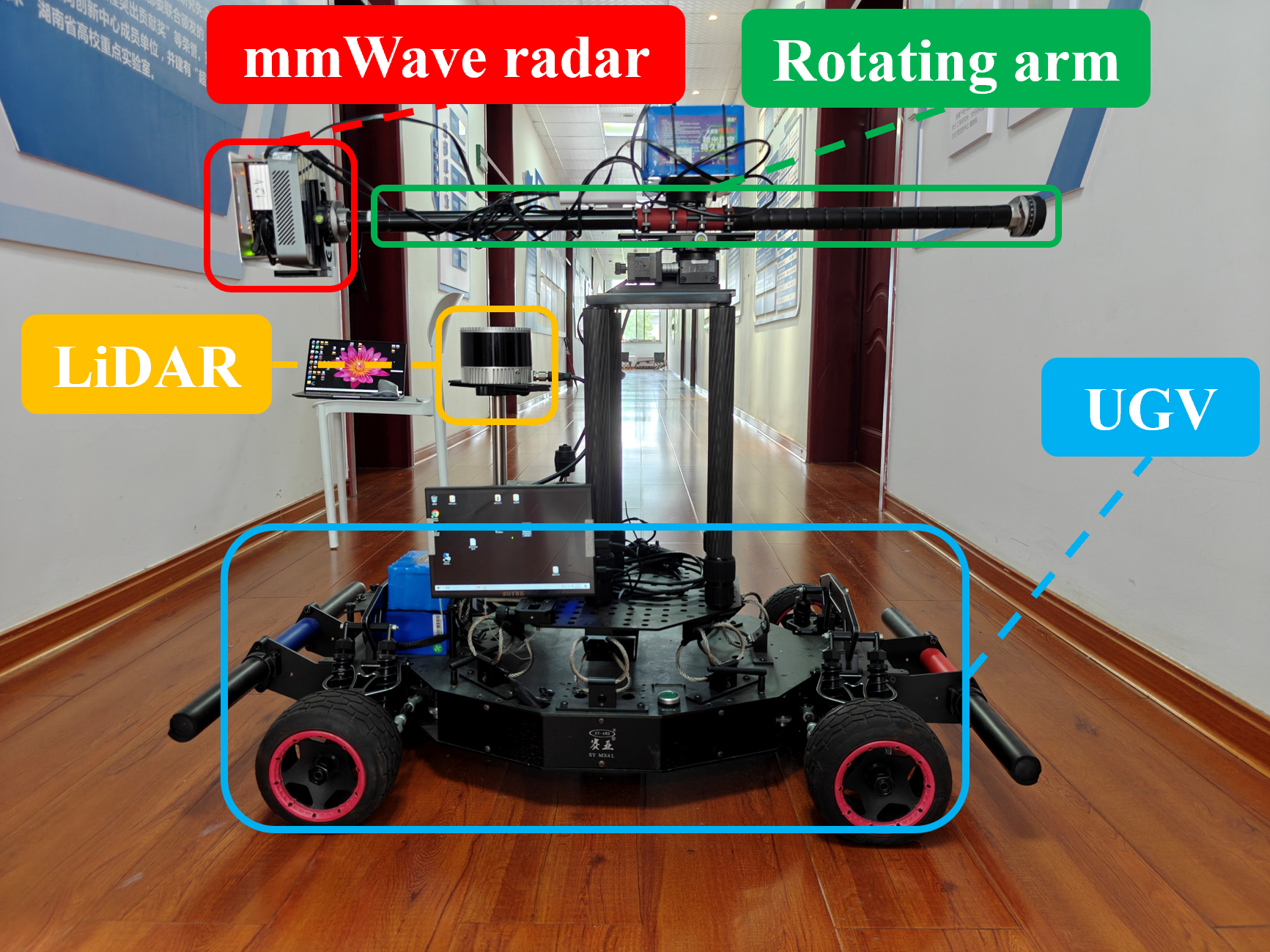}
    \caption{The Rotary Scan SAR imaging system, consisting of an array
        radar subsystem with MIMO antennas, a rotating arm subsystem, and an
        unmanned ground vehicle (UGV). The LiDAR provides auxiliary data to
        facilitate the annotation of the radar images.}
    \label{fig:radarsys}
\end{figure}

All models are trained on a single NVIDIA RTX 3090 GPU using
Adam~\cite{kingma2014adam} with cosine
annealing (300 epochs). For CARRADA and CARRADA-RAC, the loss combines weighted
cross-entropy, Dice, and coherence losses with an initial learning rate of
$1\times10^{-4}$ and batch size 3; for RS-SAR, weighted cross-entropy and Dice
losses are used with learning rate $1\times10^{-4}$ and batch size 12. Our
models take five consecutive frames as input. Evaluation follows prior work using
mIoU and mDice averaged over all classes on the test set. Full training details
and dataset descriptions are provided in the Supplementary Material.

\subsection{Frequency Selectivity Analysis}
\label{sec:freq_analysis}

\begin{figure}[t!]
    \centering
    \includegraphics[width=\linewidth]{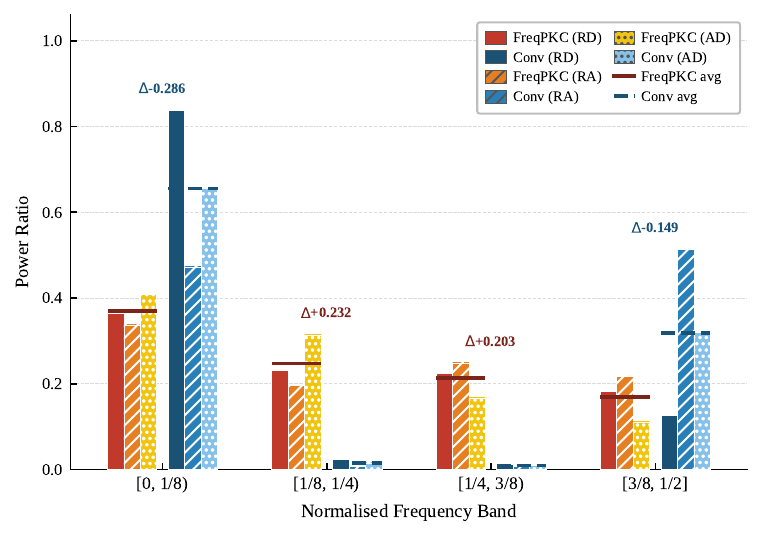}
    \caption{Output power distribution of FreqPKC and standard convolution
    layers across three radar views (RD, RA, AD) under white-noise input. The
    power ratio in each normalized band $[f_l, f_h)$ reflects the layer's
    frequency selectivity. Bold lines: group averages; $\Delta$: FreqPKC
    average minus Conv average.}
    \label{fig:band_power}
\end{figure}

To validate the frequency-domain motivation of FreqPKC, we analyze the output
frequency selectivity of FreqPKC and standard convolution layers by feeding
spectrally flat white noise into trained models and measuring the output power
ratio across normalized frequency bands. As shown in Fig.~\ref{fig:band_power},
standard convolution concentrates over $65\%$ of its output energy in the lowest
frequency band $[0, 1/8)$, reflecting its inherent low-pass filtering nature. In
contrast, FreqPKC redistributes output power into the mid-frequency bands $[1/8,
    1/4)$ and $[1/4, 3/8)$, with gains of $\Delta = +0.232$ and $\Delta = +0.203$
over standard convolution, respectively. In contrast to the low-pass nature of
standard convolution, FreqPKC better retains the structured spectral features
unique to radar targets, validating the core design of our method: the guard
bandwidth is adaptively modulated based on the spectral energy distribution,
enabling frequency-selective feature extraction well-aligned with the physical
properties of radar signals. Notably, the reinforced mid-frequency bands
$[1/8, 3/8)$ coincide with the target-background discriminative region
identified in Fig.~\ref{fig:intro_spectrum}(\subref{fig:intro_class_spectrum}), echoing the sub-band
design of Section~\ref{sec:freqpkc}.

\subsection{Ablation Studies}

We first determine the default cutoff scale factor set $\mathcal{K}$ by a
configuration sweep on CARRADA-RAC (Supplementary Table~S2). Coarse
decompositions ($\mathcal{K}=\{2\}$, $\{2,4\}$) lack the spectral resolution
to separate targets at different scales, while finer configurations
($\{2,4,6,8\}$, $\{2,4,6,8,10\}$) introduce redundant high-frequency sub-bands
whose attention maps overfit training-set spectral patterns; the four-band
decomposition $\mathcal{K}=\{2,4,8\}$ achieves the best accuracy and is
adopted in all subsequent experiments.

Beyond the band configuration, we further ablate the three FA sub-modules on
CARRADA-RAC (Table~\ref{tab:fa_ablation}): removing the frequency
decomposition (feeding the full-band feature directly into the prediction
head), freezing the band-level modulation to uniform weights, and replacing
the adaptive $B^{G\dagger}$ with a globally learnable scalar. Every variant
drops consistently across all four metrics, with the band-level modulation
($-2.7$ RD mIoU) and the adaptive bandwidth ($-2.1$) contributing most.
Moreover, all single-removal variants still outperform the FA-free PKCIn-Net
baseline by $2.8$ to $4.0$ RD mIoU, indicating that the gains arise from the
cooperation of the three components rather than parameter growth.

\begin{table}[!t]
    \centering
    \caption{Component-wise ablation of the FA module on the CARRADA-RAC dataset.}
    \label{tab:fa_ablation}
    \resizebox{\linewidth}{!}{%
        \begin{tabular}{lcccc}
            \toprule
            \multirow{2}{*}{Model} & \multicolumn{2}{c}{RD View} & \multicolumn{2}{c}{RA View} \\
            \cmidrule(lr){2-3} \cmidrule(lr){4-5}
            & mIoU & mDice & mIoU & mDice \\
            \midrule
            PKCIn-Net (w/o FA) & 60.1 & 72.6 & 47.3 & 58.7 \\
            \midrule
            w/o frequency decomposition & 64.1 & 76.0 & 49.2 & 60.9 \\
            w/o band-level modulation & 62.9 & 74.7 & 49.7 & 61.4 \\
            w/o adaptive $B^{G\dagger}$ & 63.5 & 75.3 & 49.5 & 61.1 \\
            \midrule
            \textbf{FreqPKC-Net (full FA)} & \textbf{65.6} & \textbf{77.2} & \textbf{52.3} & \textbf{64.2} \\
            \bottomrule
        \end{tabular}
    }
\end{table}

\subsection{Comparison with SoTA}

\begin{table}[!t]
    \centering
    \caption{Quantitative comparison with SoTA methods on the CARRADA dataset. The best results are highlighted in \textbf{bold}. The second best results are \underline{underlined}.}
    \label{tab:sota_carrada}
    \resizebox{\linewidth}{!}{%
        \begin{tabular}{lccccc}
            \toprule
            \multirow{2}{*}{Frameworks}                                                 &
            \multirow{2}{*}{\begin{tabular}[c]{@{}c@{}}\#Params\\ @Frames\end{tabular}} &
            \multicolumn{2}{c}{RD View}                                                 &
            \multicolumn{2}{c}{RA View}                                                                                                                                                \\
            \cmidrule(lr){3-4} \cmidrule(lr){5-6}
                                                                                        &          & mIoU               & mDice              & mIoU               & mDice              \\
            \midrule
            RAMP-CNN\cite{gao2020ramp}                                                  & 106.4M@9 & 56.6\%             & 68.5\%             & 27.9\%             & 30.5\%             \\
            MVA-Net \cite{ouaknine2021multi}                                            & 4.8M@3   & 53.5\%             & 65.3\%             & 37.1\%             & 44.8\%             \\
            TMVA-Net \cite{ouaknine2021multi}                                           & 5.6M@5   & 56.1\%             & 68.0\%             & 37.7\%             & 46.2\%             \\
            TransRadar \cite{dalbah2024transradar}                                      & 4.9M@5   & 56.6\%             & 68.9\%             & 37.9\%             & 46.8\%             \\
            \midrule
            PKCOn-Net \cite{zhang2023peakconv}                                          & 6.3M@1   & 59.5\%             & 71.3\%             & 39.5\%             & 48.3\%             \\
            PKCIn-Net \cite{zhang2023peakconv}                                          & 6.3M@5   & 60.0\%             & 72.2\%             & 42.0\%             & 52.1\%             \\
            AdaPKC$^{\xi}$-Net \cite{li2024adapkc}                                      & 6.3M@5   & 58.7\%             & 70.5\%             & 39.8\%             & 49.4\%             \\
            AdaPKC$^{\theta}$-Net \cite{li2024adapkc}                                   & 6.3M@5   & 60.2\%             & 72.1\%             & 42.7\%             & 53.3\%             \\
            AdaPKC$^{\xi}$-Net$^{FiT}$ \cite{li2024adapkc}                              & 6.3M@5   & 62.1\%             & 74.0\%             & 44.3\%             & 55.5\%             \\
            \midrule
            AXFL-Net \cite{li2025axfl}                                                  & 4.9M@5   & \underline{64.9\%} & \underline{76.6\%} & \underline{48.2\%} & \underline{60.3\%} \\
            MARSS \cite{chen2026marss}                                                  & 9.3M@5   & 63.3\%             & 75.2\%             & 47.0\%             & 58.8\%             \\
            \midrule
            \textbf{FreqPKC-Net (Ours)}                                                 & 6.4M@5   & \textbf{65.5\%}    & \textbf{77.1\%}    & \textbf{49.1\%}    & \textbf{61.1\%}    \\
            \bottomrule
        \end{tabular}
    }
\end{table}

\begin{table}[!t]
    \centering
    \caption{Quantitative comparison with SoTA methods on the CARRADA-RAC dataset. The best results are highlighted in \textbf{bold}. The second best results are \underline{underlined}.}
    \label{tab:sota_carrada-rac}
    \resizebox{\linewidth}{!}{%
        \begin{tabular}{lccccc}
            \toprule
            \multirow{2}{*}{Frameworks}                                                 &
            \multirow{2}{*}{\begin{tabular}[c]{@{}c@{}}\#Params\\ @Frames\end{tabular}} &
            \multicolumn{2}{c}{RD View}                                                 &
            \multicolumn{2}{c}{RA View}                                                                                                                                                \\
            \cmidrule(lr){3-4} \cmidrule(lr){5-6}
                                                                                        &          & mIoU               & mDice              & mIoU               & mDice              \\
            \midrule
            RAMP-CNN \cite{gao2020ramp}                                                 & 106.4M@9 & 56.5\%             & 67.5\%             & 37.9\%             & 40.5\%             \\
            MVA-Net \cite{ouaknine2021multi}                                            & 4.8M@3   & 53.6\%             & 66.3\%             & 40.1\%             & 53.9\%             \\
            TMVA-Net \cite{ouaknine2021multi}                                           & 5.6M@5   & 56.3\%             & 68.1\%             & 46.6\%             & 57.9\%             \\
            \midrule
            PKCOn-Net \cite{zhang2023peakconv}                                          & 6.3M@1   & 59.9\%             & 71.4\%             & 45.5\%             & 57.3\%             \\
            PKCIn-Net \cite{zhang2023peakconv}                                          & 6.3M@5   & 60.1\%             & 72.6\%             & 47.3\%             & 58.7\%             \\
            AdaPKC$^{\xi}$-Net \cite{li2024adapkc}                                      & 6.3M@5   & 60.1\%             & 72.0\%             & 46.1\%             & 57.2\%             \\
            AdaPKC$^{\theta}$-Net \cite{li2024adapkc}                                   & 6.3M@5   & 60.9\%             & 72.8\%             & 47.6\%             & 59.0\%             \\
            AdaPKC$^{\xi}$-Net$^{FiT}$ \cite{li2024adapkc}                              & 6.3M@5   & 60.8\%             & 72.7\%             & 48.8\%             & 60.5\%             \\
            \midrule
            AXFL-Net \cite{li2025axfl}                                                  & 4.9M@5   & \underline{64.1\%} & \underline{75.6\%} & \underline{51.4\%} & \underline{63.4\%} \\
            MARSS \cite{chen2026marss}                                                  & 9.3M@5   & 61.4\%             & 73.4\%             & 49.0\%             & 60.9\%             \\
            \midrule
            \textbf{FreqPKC-Net (Ours)}                                                 & 6.4M@5   & \textbf{65.6\%}    & \textbf{77.2\%}    & \textbf{52.3\%}    & \textbf{64.2\%}    \\
            \bottomrule
        \end{tabular}
    }
\end{table}

\begin{figure*}[!t]
    \centering
    \setlength{\tabcolsep}{1.5pt}
    \def\QualRDH{2.95cm}
    \def\QualRAH{2.55cm}

    \begin{tabular*}{\textwidth}{@{\extracolsep{\fill}}ccccccc@{}}
        \begin{minipage}[t]{0.25\textwidth}\centering
            \includegraphics[width=\linewidth,height=\QualRDH,keepaspectratio]{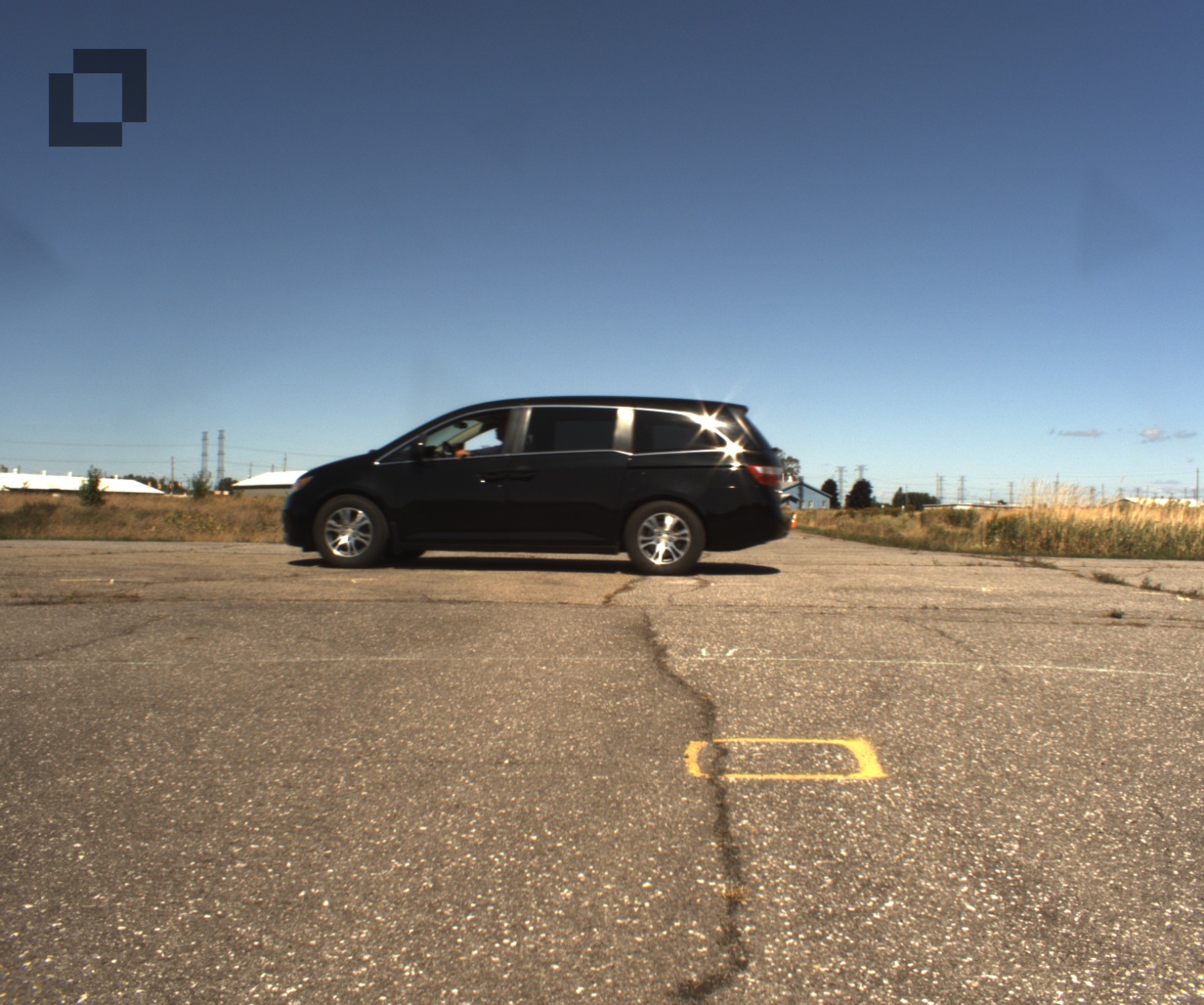}\\[-0.3em]
        \end{minipage}
        & \begin{minipage}[t]{0.08\textwidth}\centering
            \includegraphics[width=\linewidth,height=\QualRDH,keepaspectratio]{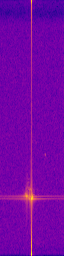}\\[-0.3em]
            \footnotesize (a)
        \end{minipage}
        & \begin{minipage}[t]{0.12\textwidth}\centering
            \includegraphics[width=\linewidth,height=\QualRDH,keepaspectratio]{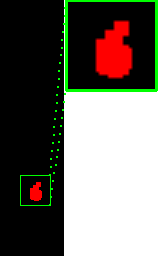}\\[-0.3em]
            \footnotesize (b)
        \end{minipage}
        & \begin{minipage}[t]{0.12\textwidth}\centering
            \includegraphics[width=\linewidth,height=\QualRDH,keepaspectratio]{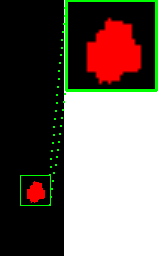}\\[-0.3em]
            \footnotesize (c)
        \end{minipage}
        & \begin{minipage}[t]{0.12\textwidth}\centering
            \includegraphics[width=\linewidth,height=\QualRDH,keepaspectratio]{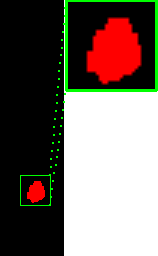}\\[-0.3em]
            \footnotesize (d)
        \end{minipage}
        & \begin{minipage}[t]{0.12\textwidth}\centering
            \includegraphics[width=\linewidth,height=\QualRDH,keepaspectratio]{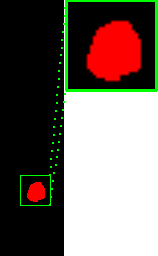}\\[-0.3em]
            \footnotesize (e)
        \end{minipage}
        & \begin{minipage}[t]{0.12\textwidth}\centering
            \includegraphics[width=\linewidth,height=\QualRDH,keepaspectratio]{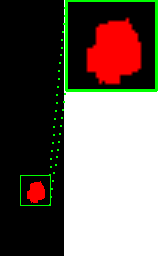}\\[-0.3em]
            \footnotesize (f)
        \end{minipage}
    \end{tabular*}

    \vspace{0.5em}

    \begin{tabular*}{\textwidth}{@{\extracolsep{\fill}}cccccc@{}}
        \parbox[t]{0.15\textwidth}{\centering\includegraphics[width=\linewidth,height=\QualRAH,keepaspectratio]{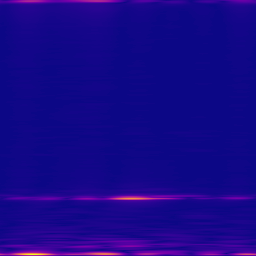}\\\footnotesize (a)}
        & \parbox[t]{0.15\textwidth}{\centering\includegraphics[width=\linewidth,height=\QualRAH,keepaspectratio]{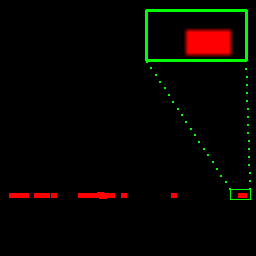}\\\footnotesize (b)}
        & \parbox[t]{0.15\textwidth}{\centering\includegraphics[width=\linewidth,height=\QualRAH,keepaspectratio]{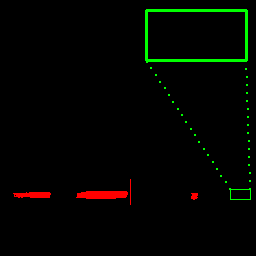}\\\footnotesize (c)}
        & \parbox[t]{0.15\textwidth}{\centering\includegraphics[width=\linewidth,height=\QualRAH,keepaspectratio]{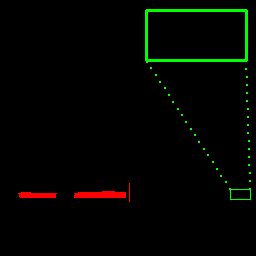}\\\footnotesize (d)}
        & \parbox[t]{0.15\textwidth}{\centering\includegraphics[width=\linewidth,height=\QualRAH,keepaspectratio]{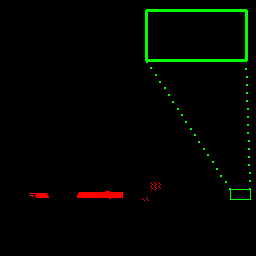}\\\footnotesize (e)}
        & \parbox[t]{0.15\textwidth}{\centering\includegraphics[width=\linewidth,height=\QualRAH,keepaspectratio]{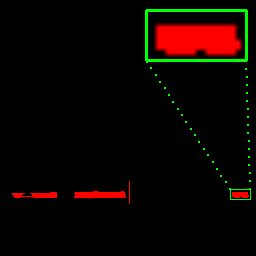}\\\footnotesize (f)}
    \end{tabular*}

    \caption{Qualitative segmentation results comparing FreqPKC-Net with selected
        baseline methods on the CARRADA-RAC dataset. One representative frame
        is shown from the test split. The \textit{top} row
        displays the camera image and RD view results, while the \textit{bottom}
        row shows the RA view results. (a) Input RD/RA tensor, (b) Ground truth
        mask, (c) PKCIn-Net, (d) AdaPKC$^{\theta}$-Net, (e)
        AdaPKC$^{\xi}$-Net$^{FiT}$, (f) FreqPKC-Net (Ours). Different colors
        represent different object categories. Black: background,
        \textcolor{red}{Red}: car, \textcolor{yellow}{Yellow}: pedestrian,
        \textcolor{cyan}{Cyan}: cyclist. }
    \label{fig:qual_carrada}
\end{figure*}

\subsubsection{Quantitative Results Analysis}
As reported in Table~\ref{tab:sota_carrada}, on CARRADA FreqPKC-Net achieves
65.5\% mIoU and 77.1\% mDice on the RD view, and 49.1\% mIoU and 61.1\% mDice
on the RA view, surpassing the second-best method AXFL-Net~\cite{li2025axfl}
by 0.6/0.5 points (mIoU/mDice) on RD and 0.9/0.8 points on RA. Results on
CARRADA-RAC (Table~\ref{tab:sota_carrada-rac}) show a similar trend, with
consistent margins over AXFL-Net.

We also compare with MARSS~\cite{chen2026marss}, a concurrent Mamba-based
method with wavelet subband decoupling. FreqPKC-Net surpasses it by 2.2/1.9
points (RD mIoU/mDice) and 2.1/2.3 points (RA) on CARRADA, and by 4.2/3.8 and
3.3/3.3 points on CARRADA-RAC, indicating that subband decoupling alone,
without adaptive modulation of the receptive field, leaves accuracy gains
untapped.

On CARRADA, the RA view benefits more from frequency-aware modulation than the
RD view (+0.9/+0.8 mIoU/mDice \textit{vs.} +0.6/+0.5), consistent with the RA
view's greater angular ambiguity and denser clutter. On CARRADA-RAC,
FreqPKC-Net maintains consistent gains on both views. This view-dependent
behavior confirms that the FA module acts as a complementary mechanism to
spatial feature learning.

We attribute the consistent improvements to two factors. First, the FA module
establishes a physically interpretable mapping from multi-band spectral energy
to guard bandwidth offsets: unlike the spatial regression in AdaPKC and
AXFL-Net, FreqPKC-Net infers directional guard bandwidths directly from
frequency-domain statistics, aligning the operator's response with the
spectral characteristics of targets and clutter (Supplementary Fig.~S4
confirms that target regions receive consistently smaller $b_{\rm eff}$).
Second, the cascaded residual decoupling with $\mathcal{K}=\{2,4,8\}$
(Supplementary Table~S2) enables band-level modulation that enhances
task-relevant components while suppressing noise-dominated bands, an explicit
multi-scale frequency awareness absent in prior methods.

The gains also widen on CARRADA-RAC: the RD margin over AXFL-Net grows from
0.6 to 1.5 mIoU, and the margins over AdaPKC$^{\xi}$-Net$^{FiT}$ reach 4.8
mIoU on RD and 3.5 on RA. Relative to the PKCIn-Net backbone, the FA module
contributes 5.5 mIoU on RD for both datasets and 7.1/5.0 mIoU on RA
(CARRADA/CARRADA-RAC). Since CARRADA-RAC refines CARRADA's RA annotations
without changing the sensor, the sustained margins across the two annotation
regimes indicate that spectral guidance is robust to label granularity and
generalizes more stably than spatial-only bandwidth regression.

On the hardest setting, the CARRADA RA view, every method degrades sharply:
AXFL-Net drops from 64.9\% to 48.2\% mIoU and AdaPKC$^{\xi}$-Net$^{FiT}$ from
62.1\% to 44.3\%, while FreqPKC-Net retains 49.1\%, widening its margin over
AdaPKC$^{\xi}$-Net$^{FiT}$ to 4.8 mIoU on this view. These gains align with
the mid-frequency bands reinforced in Section~IV-B.

\textbf{Efficiency.} We benchmark inference efficiency on a single RTX 3090
GPU (batch size 1, FP32). FreqPKC-Net adds only $0.17$M parameters ($+2.7\%$)
and $1.95$ GFLOPs ($+1.8\%$) over PKCIn-Net while delivering the best accuracy
in Tables~\ref{tab:sota_carrada} and~\ref{tab:sota_carrada-rac}; the full
comparison, including FPS and latency, is provided in the Supplementary
Material (Table~S1).

\subsubsection{Qualitative Results Analysis}
\label{sec:qualitative}
Fig.~\ref{fig:qual_carrada} shows a representative vehicle scene from the
CARRADA-RAC test split. We compare FreqPKC-Net against PKCIn-Net,
AdaPKC$^{\theta}$-Net, and AdaPKC$^{\xi}$-Net$^{FiT}$ under both RD and RA views
(AXFL-Net is excluded as its open-source implementation is not publicly
available); each subfigure shows a zoomed-in view of the corresponding region.
All methods perform comparably on
the RD view, where noise is mostly concentrated near the zero-frequency line. On
the more challenging RA view, the vehicle appears as an elongated strip, and all
comparison models suffer from target incompleteness. FreqPKC-Net produces a
segmentation mask with complete contours, better continuity, and fewer false
positives. Additional qualitative results on pedestrian and multi-object scenes
are provided in the Supplementary Material. These observations are consistent
with the quantitative gains in Tables~\ref{tab:sota_carrada}
and~\ref{tab:sota_carrada-rac}, and corroborate that explicit frequency-domain
guidance enables more physically grounded receptive field modulation than
spatial-only adaptive approaches.

\subsection{RSS Performance on RS-SAR Dataset}
\label{sec:rs_sar_results}

As reported in Table~\ref{tab:sota_rs_sar}, FreqPKC-Net achieves 42.16\%
mIoU and 49.10\% mDice on the RS-SAR dataset, outperforming all competing
methods by clear margins. Compared with the second-best model
AdaPKC$^{\theta}$-Net, FreqPKC-Net yields improvements of $+1.20$ points in
mIoU and $+2.06$ points in mDice, with only a marginal increase in parameters
(1.26M \textit{vs.}~1.25M). Notably, the mDice gain ($+2.06$ points) is
substantially larger than the mIoU gain, suggesting that the frequency-aware
design particularly improves boundary alignment on the sparser and noisier
RS-SAR data. Qualitative segmentation results on two representative RS-SAR scenes,
including a detailed per-scene analysis, are provided in the Supplementary
Material (Fig.~S6). Consistent with the quantitative results in
Table~\ref{tab:sota_rs_sar}, FreqPKC-Net achieves more accurate segmentation
of targets.

Finally, the three benchmarks involve different radar configurations: a
forward-facing 77\,GHz FMCW radar for CARRADA/CARRADA-RAC versus a
360$^\circ$ rotary-scan SAR system for RS-SAR. The consistent gains across all
three datasets suggest that the frequency-aware mechanism is not tied to a
specific sensor or annotation protocol, as the spectral separability between
targets and clutter stems from scattering physics rather than acquisition
geometry. Nevertheless, since radar spectral statistics depend on the
sensor's operating frequency, bandwidth, and resolution, $\mathcal{K}$ may
require recalibration on substantially different configurations; a systematic
cross-sensor transfer study is left as future work.

\begin{table}[!t]
    \centering
    \caption{Quantitative comparison with SoTA methods on the RS-SAR dataset. The best results are highlighted in \textbf{bold}. The second best results are \underline{underlined}.}
    \label{tab:sota_rs_sar}
    \resizebox{0.8\linewidth}{!}{%
        \begin{tabular}{lccc}
            \toprule
            Frameworks                                & \begin{tabular}[c]{@{}c@{}}\#Params\\ @Frames\end{tabular} & mIoU                & mDice               \\
            \midrule
            MVA-Net \cite{ouaknine2021multi}          & 1.30M@3                                                    & 39.72\%             & 45.19\%             \\
            TMVA-Net \cite{ouaknine2021multi}         & 1.20M@5                                                    & 39.44\%             & 44.61\%             \\
            \midrule
            PKCIn-Net \cite{zhang2023peakconv}        & 1.25M@5                                                    & 39.80\%             & 45.13\%             \\
            AdaPKC$^{\xi}$-Net \cite{li2024adapkc}    & 1.25M@5                                                    & 40.23\%             & 45.94\%             \\
            AdaPKC$^{\theta}$-Net \cite{li2024adapkc} & 1.25M@5                                                    & \underline{40.96\%} & \underline{47.04\%} \\
            \midrule
            \textbf{FreqPKC-Net (Ours)}               & 1.26M@5                                                    & \textbf{42.16\%}    & \textbf{49.10\%}    \\
            \bottomrule
        \end{tabular}
    }
\end{table}

\section{Conclusion}
\label{sec:conclusion}
Existing radar semantic segmentation methods overlook the relationship between
PRF design and spectral characteristics. We reveal that the guard bandwidth in
PKC directly governs its frequency passband, and propose Frequency-Aware
PeakConv (FreqPKC), a radar-specific neural operator that leverages spectral
energy distribution to guide adaptive PRF modulation. The FA module decomposes
features via cascaded residual decoupling, applies band-level learnable
modulation, and maps multi-band spectral statistics into directional guard
bandwidth offsets, enabling the PRF to balance high-frequency target detail
against low-frequency clutter in a physically interpretable manner. Building on
FreqPKC, we develop FreqPKC-Net, a multi-view RSS framework with cross-view
fusion. We also construct RS-SAR, a high-resolution Rotary Scan SAR dataset
with full 360$^\circ$ panoramic coverage, complementing existing benchmarks with
a distinct sensor modality and an order-of-magnitude higher spatial resolution.
Experiments on CARRADA, CARRADA-RAC, and RS-SAR show that FreqPKC-Net
consistently outperforms SoTA methods, and the multi-dataset evaluations confirm its
strong generalization across different radar sensors.


\bibliographystyle{IEEEtran}
\bibliography{Ref}

\end{document}